\documentclass[aos, preprint, reqno]{imsart}

\RequirePackage[OT1]{fontenc}
\RequirePackage{amsthm,amsmath}
\RequirePackage[numbers]{natbib}
\RequirePackage[colorlinks,citecolor=blue,urlcolor=blue]{hyperref}

\startlocaldefs
\numberwithin{equation}{section}
\endlocaldefs

\usepackage{amsfonts}
\usepackage{amsmath}
\usepackage{amsthm}

\usepackage{hongwei}
\usepackage{hongwei_2}

\usepackage{graphicx} 
\usepackage{epsfig} 
\usepackage{subfigure}

\usepackage{algorithm}
\usepackage{algorithmic}

\begin{document}

\begin{frontmatter}
\title{Error Rate Bounds in Crowdsourcing Models}
\runtitle{Error Rate Bounds in Crowdsourcing Models}
\thankstext{T1}{ Email: \texttt{hwli@stat.berkeley.edu} (Hongwei Li),  \texttt{binyu@stat.berkeley.edu} (Bin Yu) and \texttt{dengyong.zhou@microsoft.com} (Dengyong Zhou)}



\begin{aug}
\author{\fnms{Hongwei} \snm{Li}$^1$\ead[label=e1]{hwli@stat.berkeley.edu}},
\author{\fnms{Bin} \snm{Yu}$^{1,2}$\ead[label=e2]{binyu@stat.berkeley.edu}}
\and
\author{\fnms{Dengyong} \snm{Zhou}$^3$
\ead[label=e3]{dengyong.zhou@microsoft.com}
\ead[label=u1,url]{}}

\runauthor{Li, Yu and Zhou}

\affiliation{Department of Statistics, UC Berkeley $^1$,  \\ Department of EECS, UC Berkeley $^2$  and \\ Microsoft Research, Redmond $^3$}

\address{367 Evans Hall, \\Department of Statistics\\
 University of California\\
Berkeley, CA 94720-1776, USA\\
\printead{e1}\\
\phantom{E-mail:\ }\printead*{e2}}

\address{ Microsoft Research\\
One Microsoft Way\\
Redmond, WA 98052, USA \\
\printead{e3}\\
}
\end{aug}

\begin{abstract}

Crowdsourcing is an effective tool for human-powered computation on many tasks challenging for computers. In this paper, we provide finite-sample exponential bounds on the error rate (in probability and in expectation) of hyperplane binary labeling rules under the Dawid-Skene crowdsourcing model. The bounds can be applied to analyze many common prediction methods, including the majority voting and weighted majority voting.
These bound results could be useful for controlling the error rate and designing better algorithms.  We show that the oracle Maximum A Posterior (MAP) rule approximately optimizes our upper bound on the mean error rate for any hyperplane binary labeling rule, and propose a simple data-driven weighted majority voting (WMV) rule (called one-step WMV) that attempts to approximate the oracle MAP and has a provable theoretical guarantee on the error rate.
Moreover, we use simulated and real data to demonstrate that the
data-driven EM-MAP rule is a good approximation to the oracle MAP rule, and to demonstrate that the mean error rate of the data-driven EM-MAP rule is also bounded by the mean error rate bound of the oracle MAP rule with estimated parameters plugging into the bound.

\end{abstract}



\end{frontmatter}


\section{Introduction}

There are many tasks that can be easily carried out by people but tend to be hard for computers, e.g. image annotation and visual design.
When these tasks require  large scale data processing,
outsourcing them to experts or well-trained people may be too expensive.
Crowdsourcing has recently emerged as a powerful alternative.
It outsources tasks to a distributed group of people (usually called workers) who might be inexperienced in these tasks. However, if we can appropriately aggregate the  outputs from a crowd,  the aggregated results could be as good as the results by an expert
\cite{Dawid_JRSS79,Liu2012,Raykar_jmlr10, Smyth1995,Snow_emnlp08, Welinder_nips10,Whitehill_nips09,Yan_icml10,Zhou2012}.

The flaws of crowdsourcing are apparent. Each worker is paid purely based on how many tasks that he/she has completed (for example, one cent for labeling one image).  No ground truth is available to evaluate how well he/she has performed in the tasks. So some workers may randomly submit answers independent of the questions when the tasks assigned to them are beyond their expertise. Moreover, workers are usually not persistent. Some workers may complete many tasks, while the others may only  finish very few tasks even just one.

In spite of these drawbacks, is it still possible to get reliable answers in a crowdsourcing system? The answer is yes. In fact, majority voting (MV)  has been able to generate fairly reasonable results \cite{Liu2012, Raykar_jmlr10,Snow_emnlp08,Zhou2012}. However, majority voting treats each worker's result as equal in quality. It does not distinguish a spammer from a diligent worker. So we can expect that majority voting can be significantly improved upon.

The first improvement over majority voting might date back to Dawid and Skene \cite{Dawid_JRSS79}. They assumed that each worker is associated with an unknown confusion matrix. Each off-diagonal element represents misclassification rate from one class to the other, while the diagonal elements represent the accuracy in each class. According to the observed labels by the workers, the maximum likelihood principle is applied to jointly estimate unobserved true labels and  worker confusion matrices. The likelihood function is non-convex. However, a local optimum can be obtained by using the Expectation-Maximization (EM) algorithm, which can be initialized by majority voting.

Dawid and Skene\rq{}s approach \cite{Dawid_JRSS79} can be straight-forwardly  extended by assuming true labels to be generated from a logistic model  \cite{Raykar_jmlr10} or putting prior over worker confusion matrices \cite{Liu2012, Raykar_jmlr10}. One may simplify the assumption made in \cite{Dawid_JRSS79} to consider a symmetric confusion matrix \cite{Karger_NIPS2011, Raykar_jmlr10}, which we call  the \sds  model.

Recently, significant efforts have been made to analyze error rate for the algorithms in the literature.
In \cite{Karger_NIPS2011}, Karger et al. provided asymptotic error bounds for their iterative algorithm and  also majority voting. However, the error bound for their specific iterative algorithm cannot be generalized to other prediction rules in crowdsourcing and the  asymptotic bounds may not be that practical since we always only have finite number of tasks. Additionally, their results depend on the assumption that the same number of items were assigned to each worker, and the same number of workers labeled each item.
According to the analysis in \cite{Liu2012} by Liu et al., this assumption is restrictive in practical case.
Ho et al. \cite{Ho2013} formulated the adaptive task assignment problem, which considers adaptively assigning workers to different types of tasks according to their performance, into an optimization problem, and provided  performance guarantee of their algorithm.  Meanwhile, they provided a mean error rate bound for weighted majority voting, which can be viewed as a special case of our general bound of mean error rate for hyperplane rules.



In this paper, we focus on providing bounds on the error rate under crowdsourcing models of which the effectiveness on real data has been evaluated in \cite{Dawid_JRSS79,Liu2012, Raykar_jmlr10}. Our main contributions are as follows. We derive error rate bounds in probability and in expectation for a finite number of workers and instances
under the Dawid-Skene  model (with the \sds  model as a special case).
Moreover, we provide error bounds for the oracle Maximum A Posterior (MAP) rule and a data-driven weighted majority voting (WMV), and show that the oracle MAP rule approximately optimizes the upper bound on the mean error rate for any hyperplane rule.
Under the \sds  model, we use simulation to demonstrate that the data-driven EM-MAP rule approximates  the oracle MAP rule well.
To the best of our knowledge, this is the first work which focuses on the error rate analysis on general prediction rules under the practical \ds model for crowdsourcing, which can be used for analyzing error rate and sample complexity of  algorithms like those in \cite{Ho2013, Karger_NIPS2011}.

\section{Problem setting and formulation}\label{sec:problem_formulate}

We focus on binary labeling in this paper. Assume that a set of workers are assigned to label certain items that are available on the Internet. For instance, whether an image of an animal is that of a cat or a dog, or if a face image is male or female.

Formally, suppose we have $M$ workers, and $N$ items. For convenience, we denote $\M= \hua{1, \cdots, M}$ and $\N=\hua{1,\cdots, N}$.
The label matrix is denoted by $Z\in \hua{\pm 1, 0}^{M\times N}$, in which $Z_{ij}$ is the label of the $j$-th item given by the $i$-th worker. It will be $0$ if the corresponding label is missing.
Throughout the paper, we use $\yj$ as the true label for $j$-th item, and $\hyj$ as the predicted label for the $j$-th item by an algorithm. At the same time, any parameter with a hat $\hat{~}$ is an estimate for this parameter.
Let $\pri= \P(\yjp1)$ for any $ j\in [N]$ denote the prevalence of label ``$+$" in the true labels of the items.
We introduce the indicator matrix $T=(\Tij)_{M\times N}$, where $\Tij=1$ indicates that entry $(i,j)$ is observed, i.e., the $i$-th worker has labeled the $j$-th item, and $\Tij=0$ indicates entry $(i,j)$ is unobserved.
 Note that $T$ and $Z$ are observed together in our crowdsourcing setting, and both the number of items labeled by each worker and the number of workers assigned to each item are random. 
The sampling probability matrix is denoted by $\Q= (\qij)_{M\times N}$, where $\qij= \P(\Tij=1)$, i.e., $\qij$ is the probability that the $i$th worker labels the $j$th item.
When $\qij= \qi\in(0,1], \forall i\in\M, j\in\N$, we call it \emph{ sampling with probability vector $\qvec=(\q_1, \cdots, \q_M)$}. If $\qij=\qs \in(0,1], \forall i\in\M, j\in\N$, then we call it \emph{sampling with constant probability $\q$.} 


We will discuss two models that are widely used for modeling the quality of the workers\cite{Karger_NIPS2011, Liu2012, Raykar_jmlr10, Zhou2012}. They were first proposed by Dawid and Skene \cite{Dawid_JRSS79}:

\hwem{\ds model.~}
We distinguish the accuracy of workers on the positive class and the negative class. Some workers might work better on labeling the items with true label ``$+$", and some might work better at labeling the items with  true label ``$-$".
The true positive rate (sensitivity) and the true negative rate (specificity) are denoted as follows respectively: for $ i=1,2, \cdots, M$

\begin{eqnarray}
\qquad 
\ppi := 	P(\zijp1|\yjp1, \Tij=1)
\connect
\pni :=	P(\zijn1| \yjn1, \Tij=1). 
\end{eqnarray}

Then the parameter set will be
$
\para= \hua{\hua{\ppi,\pni}_{i=1}^M, \Q, \pi}
$ under this model.

\hwem{\sds model.~}
 We assume that the $i$-th worker labels the item correctly with a fixed probability $\wi=\P(\zij=y_j | y_j, \Tij=1)$, i.e., $\ppi=\pni=\wi$.
In this case, no matter whether an item is from positive class or negative class, the worker labels it with the same accuracy.
Therefore, the parameter set is
$
\para= \hua{\hua{w_i}_{i=1}^M, \Q, \pi}.
$


Under both models above, the posterior probability of the label for each item to be ``$+$" is defined as:
$
\rhoj= \P( \yjp1 |Z,T,\para), \quad \forall j\in [N]. \label{def:rho}
$
Given an estimation or a prediction rule, suppose that its predicted label for item $j$ is $\hyj$, then our objective is to minimize the error rate.
Since the error rate is random, we are also interested in its expected value (i.e., the \emph{mean error rate}). Formally, error rate and its expected value are:
\vspace{-2mm}
\begin{eqnarray}
\errRate = \inv{N}\sumj \I{\hyj\neq y_j}
\quad\connect\quad
\E[\errRate]= \inv{N}\sumj \P(\hyj \neq y_j).
\end{eqnarray}
\vspace{-5mm}

The rest of the paper is organized as follows. In Section \ref{sec:pac}, we present a finite-sample bound on the error rate of a hyperplane rule in probability and also in expectation under the \ds model. In Section \ref{sec:MLE}, we apply our analysis to the label inference by Maximum Likelihood method, illustrate the bound on the oracle MAP rule and present a bound of a simple data-driven weighted majority voting under the \sds model. Experimental results on simulated and real-world dataset are presented in  section \ref{sec:experiment}.
Note that the proofs are deferred to the supplementary materials.


\section{Error rate bounds} \label{sec:pac}

In this section, we provide finite-sample bounds on error rate of any hyperplane rule under the Dawid-Skene model in high probability and in expectation.


\subsection{Bounds on the error rate in high probability }

A hyperplane prediction or estimation rule is a rectified linear function of the observation matrix $Z$ in a high dimensional space: given an unnormalized weight vector $\vweight\in\R^M$ (independent of $Z$) and an shift constant $\aconst$, for the $j$th item, the rule estimates its label as
$
\hyj= \sign(\sumi \vi \zij + \aconst).
$
In the rest of this paper, we call it \emph{hyperplane rule}. It is a very general rule with special cases including majority voting (MV) which has $\vi=1$ for all $i$ and $\aconst=0$.

Next we present two general theorems to provide finite-sample error rate bounds for hyperplane rules under the Dawid-Skene model. Before that, we introduce some notations as follows:
\begin{eqnarray}
&& \eiplus= \kua{\sumi \qij\vi(2\ppi-1) + \aconst} \connect
 \eiminus = \kua{\sumi \qij\vi(2\pni-1) - \aconst} \\
&& \tone = \minj \frac{ \eiplus \minop \eiminus  }{\vnorm}
\connect
 \ttwo = \maxj \frac{ \eiplus \maxop \eiminus } {\vnorm}
\nonumber \\
&& \phi(x)= e^{-\frac{x^2}{2}}\qquad x\in\R
\connect
 \D(x||y)= x\ln{\frac{x}{y}} + (1-x)\ln{\frac{1-x}{1-y}} \quad, \;  x,y\in(0,1) \nonumber ,
\end{eqnarray}
where $||\cdot||_2$ is $L_2$ norm,  $x \wedge y$ is $\min\hua{x,y}$ and  $x\vee y$ is $\max\hua{x,y}$.


\begin{thm} \label{thm:mainExp}

For a given sampling probability matrix $\Q = (\qij)_{M\times N}$, let $\hyj$  be the estimate by the hyperplane rule with weight vector $\vweight$ and shift constant $\aconst$.
For any $\epsilon \in (0,1)$, we have

(1) when \quad
$
\tone \geq \boundTone,
$
\qquad~~~
$
\PErrExp ;
$

(2) when \quad
$
\ttwo \leq \boundTtwo,
$
\qquad
$
\PErrExpRev .
$

\end{thm}

\remark  In fact, we have $\tone\vnorm\leq \E\braket{\sumi\vi\zij+a} \leq \ttwo\vnorm, \forall j\in\N$, thus  $\tone$ and $\ttwo$ are two very important quantities for controlling error rate if a hyperplane rule with fixed weights.
Note that for a fixed sampling probability matrix, if the weights are positive, then the better the worker over random guessing (or the bigger $2p_i^+ -1$) for ````$+$"" labels and the larger the shift $a$, the larger the $\eiplus$. Similarly we can interpret $\eiminus$.
Usually we are free to choose $\vweight$ and $\aconst$, and in some situation we can also control $\Q$, so the most important factors that we cannot control are $\ppi$ and $\pni$.

To control the probability of error rate exceeding $[-\epsilon, \epsilon]$ by $\delta$, we have to solve the equation $\exp\hua{-ND(\epsilon||\phi(\tone))} = \delta$, which cannot be solved analytically, so we need to consider about a method which can tell us what's the minimum $\tone$ for bounding the error rate with probability at least $1-\delta$. The next theorem serves this purpose.
For notation convenience, we define two constants $C$ and $G$ for $\epsilon, \delta \in (0, 1)$:
\begin{eqnarray}
C (\epsilon, \delta) =\constC
\connect
G (\epsilon, \delta) = \constG, \label{def:G} \quad 
\end{eqnarray}
where $\H(\epsilon)= -\epsilon\ln\epsilon -(1-\epsilon)\ln(1-\epsilon)$.

\begin{thm} \label{thm:general}

Under the same setting as in Theorem \ref{thm:mainExp}, for $\forall \epsilon,\delta\in (0,1)$, we have
\\
(1) if
$
\tone \geq \sqrt{2\ln C (\epsilon, \delta)},
$
then \quad
$
\PErrorBound.
$
\\
(2) If
$
\ttwo \leq -\sqrt{2\ln G (\epsilon, \delta)},
$
then\quad
$
\PErrorReverse.
$
\end{thm}

To gain insights, we consider a simple and common method -- majority voting (MV) with constant probability $\qs$ sampling entries under the \sds model, i.e., $\ppi=\pni=\wi, \forall i\in\M$. In this case, the weight of each worker is the same.  The  result below follows from Theorem \ref{thm:mainExp} by taking $\qij=q$, $\ppi=\pni=\wi$, $\vweight_i=1$ and $\aconst=0$.

\begin{cor}

Under the \sds model, for majority voting with constant
probability sampling $\qs \in (0,1]$, if
$
\wbar \geq \inv{2} + \inv{q} \sqrt{\inv{2M}\constOne},
$
then
$
\PError \geq 1 -e^{-N \D(\epsilon||\psi)},
$
where $\psi= e^{-2M\qs^2(\wbar - 0.5)^2}$.
\end{cor}

\remark This result implies that for the error rate to be small, the average accuracy of workers $\wbar$ has to be better than random guessing by $\Omega(q^{-1}M^{-0.5})$. This requirement will be easier to satisfy with more workers(larger $M$), and each worker labels more items ($q$ close to 1).

\subsection{Bounds on the error rate in expectation}\label{sec:mer_bound}

One is often interested in bounding the mean error rate for a general hyperplane rule, since the mean error rate gives the expected proportion of items wrongly labeled.

\begin{thm}\label{thm:bound_meanErrorRate}
\hwem{(Mean error rate bounds under the \ds model) }
Under the same setting as in Theorem \ref{thm:mainExp}, with $\cH$ and $\sigtwo$ defined as follows
\begin{eqnarray}\label{def:cAndSigma2}
\cH = \frac{\Linfnorm{\vweight}}{\Ltwonorm{\vweight}}
\text{~~and~~}
\sigtwo =\maxj \frac{1}{\vnorm^2}\braket{\kua{ \sumi \vi^2\qij\kua{1- \qij(2\ppi-1)^2}} \maxop \kua{ \sumi\vi^2\qij\kua{1-\qij(2\pni-1)^2}}},~
\end{eqnarray}
\\
(1)
if  \quad $\tone \geq 0$, \quad then \quad
$
\MER \leq \min \hua{ \exp\kua{ - {\tone^2 \over 2}},~ \exp\kua{- \frac{\tone^2}{2\kua{\sigtwo + \cH\tone/3}} } };
$
\\
(2)
if \quad $\ttwo \leq 0$, \quad then
$
\MER \geq 1- \min\hua{\exp\kua{-{\ttwo^2\over 2}},~ \exp\kua{- \frac{\ttwo^2}{2\kua{\sigtwo - \cH\ttwo/3}} } }.
$

\end{thm}

\remark In fact, $\cH\vnorm$ is an upper bound on $|\vi\zij|, \forall i\in\M, j\in\N$, and $\sigtwo\vnorm^2$ is an upper bound on the variance of $\kua{\sumi\vi\zij+a}$, which is rectified to predict the label of the $j$th item.



The next corollary covers the mean error rate bounds under the \sds model, which can be derived from last theorem by letting $\ppi=\pni=\wi$ and enlarging $\sigtwo$ for clarity.

\begin{cor}\label{res:mer_onecoin}
Under the \sds model, for a given sampling probability matrix $\Q = (\qij)_{M\times N}$, assuming the prediction rule is a hyperplane rule with weight $\vweight$ and shift constant $\aconst$, then with $\cH$ defined as in (\ref{def:cAndSigma2}) 
and $\sigtwopri = \maxj \frac{1}{\vnorm^2}{\kua{ \sumi \vi^2\qij}}$,
\vspace{-2mm}
\begin{eqnarray}
 \tonepri \= \min_{j\in \N} \inv{\vnorm}\kua{\sumi \qij \vi(2w_i-1) - |\aconst|}
\connect
\ttwopri= \max_{j\in \N} \inv{\vnorm}\kua{\sumi \qij \vi(2w_i-1) + |\aconst|},  \qquad~~
\end{eqnarray}
\vspace{-2mm}
\\
(1)
if  \quad $\tonepri \geq 0$, \quad then \quad
$
\MER \leq \min \hua{ \exp\kua{ - {\tonepri^2 \over 2}},~ \exp\kua{- \frac{\tonepri^2}{2\kua{\sigtwopri + \cH\tonepri/3}} } };
$
\\
(2)
if \quad $\ttwopri \leq 0$, \quad then
$
\MER \geq 1- \min\hua{\exp\kua{-{\ttwopri^2\over 2}},~ \exp\kua{- \frac{\ttwopri^2}{2\kua{\sigtwopri - \cH\ttwopri/3}} } }.
$

\end{cor}

Due to the complicated forms, the results above might not be very intuitive. Let us look at the majority voting case by applying $\vi= 1$ for all $i$ and $\aconst=0$ to the first part of bound in Corollary \ref{res:mer_onecoin}.

\begin{cor}\label{res:oneCoin_MER}

For the majority voting under the \sds model and constant probability sampling $\qs \in(0,1]$,
\\
(1) if \qquad $\wbar > \inv{2}$, \quad then\quad
$
\MER \leq e^{-2Mq^2(\wbar -\inv{2})^2};
$
\\
(2) if \qquad $\wbar < \inv{2}$,   \quad then\quad
$
\MER \geq 1 - e^{-2Mq^2(\wbar -\inv{2})^2}.
$

\end{cor}

\remark  The mean error rate of MV will exponentially decay with $M$ increase if the average accuracy of labeling by the workers is better than random guessing, and the gap between the average accuracy and 0.5 plays an important role in the bound.
 In particular, it implies that
(1) if $\limM{\wbar} > \inv{2}$,  then  $\limM{\kua{\MER}} = 0$; and
(2) if $\limM{\wbar} < \inv{2}$,  then $\limM{\kua{\MER}} = 1.$

As mentioned earlier, a main result in \cite{Karger_NIPS2011} is very similar to our result in Corollary \ref{res:oneCoin_MER}. In order to compare, we rewrite their results with our notations, and the upper bound of MV in \cite{Karger_NIPS2011} (page 5, (3))  will be $\exp\kua{- Mq(\wbar - 0.5)^2}$.  It is different from ours by $2\qs$ in the exponents. If $q>0.5$, our bound will be tighter.  It is worthy of mentioning that the results in \cite{Karger_NIPS2011} is asymptotic (with $N\rightarrow\infty$), while ours here is applicable to both asymptotic and finite sample situation. And they assumed the number of items labeled by each worker is the same and the number of workers assigned to each item is also the same, while we did not make that assumption.



\section{Data-driven EM-MAP rule and  one-step weighted majority voting}\label{sec:MLE}

If we know the posterior probability $\rhoj$ (defined in section \ref{sec:problem_formulate}) of the label of each item, then the Bayesian classifier predict $\hyj = 2\I{\rhoj > 0.5} -1$.
Thus if we estimate the posterior $\rhoj$ well, we can apply the same rule to predict the true label with the estimated posterior probability.
One natural way to approach it is to apply the Maximum Likelihood method to the observed label matrix in order to estimate the parameter set $\Theta$ and consequently the posterior.

\subsection{Maximum A Posteriori (MAP) rule and the oracle MAP rule}

As in  \cite{Dawid_JRSS79}, we can apply the EM algorithm to obtain
the maximum likelihood estimate for the parameters and the posterior $\hrj$. With $\hrj$, each item can be assigned with the label which has the largest posterior, that is, the prediction function with MAP rule is
$
\hyj = 2\I{\hat{\rhoj} > 0.5}-1,
$
where $\hat{\rhoj}$ is the estimated posterior probability. We call the method above the \emph{ EM-MAP rule}.

However, the EM algorithm cannot guarantee convergence to the global maximum of the likelihood function.
The estimated parameters might not be close to the true parameters and similarly for the estimated posterior. It is known that EM algorithm is generally hard to conduct error rate analysis due to its iterative nature.
Nevertheless, we can consider the \emph{oracle MAP rule}, which knows the true parameters and thus uses the true posterior $\rhoj$ in MAP rule to label items, i.e.,
$
\hyj= 2\I{\rhoj > 0.5}-1.
$
 We can apply the mean error rate bounds from section \ref{sec:mer_bound} to the oracle MAP rule.

\begin{thm}\label{res:msr_oracleMAP}

For the oracle MAP rule knowing the true parameters $\para=\hua{\hua{\wi}_{i=1}^M, \Q, \pi}$, its prediction function is $\hyj= 2\I{\rhoj > 0.5}-1$, then under the \sds model, the  oracle MAP rule is a hyperplane rule, i.e.,
$
\hyj= \sign\kua{\sumi\vi\zij + \aconst} \text{ with } \vi= \ln\upratio{\wi} \text{ and } \aconst= \ln\upratio{\pi},
$
thus the same mean error rate bounds and their conditions as in Corollary \ref{res:mer_onecoin} also hold here.

\end{thm}

\remark Although it is hard to obtain performance guarantee for the EM-MAP rule, empirically it  has almost the same performance as the oracle MAP rule in simulation when $\wbar>0.5$, which will be shown in Section \ref{sec:experiment}. This suggests that the bound on the mean error rate of the oracle MAP rule could be useful for estimating the error rate of the EM-MAP rule in practice as we will do in Sec. \ref{sec:experiment}.

\subsection{The oracle MAP rule and oracle bound-optimal rule} \label{sec:bound_optimal}

In this section, we will explore the relationship between the oracle MAP rule and the error rate bound under the \sds model for clarity. Meanwhile, we consider the situation that the entries in the observed label matrix are sampled with a constant probability $\qs$  for simplicity. 

Let us look closely at the mean error rate bound in Corollary \ref{res:mer_onecoin}.(1). When sampling with a constant probability $\qs$, the bound is monotonously decreasing w.r.t. $\tonepri$ on [0, $\infty$) and $\sigtwopri=\qs$, so optimizing the upper bound is equivalent to maximizing $\tonepri$,
\vspace{-2mm}
\begin{eqnarray}
(\vstar, \astar)&=&  \argmax_{\vweight \in \R^M, \aconst\in \R} \tonepri
=   \argmax_{\vweight \in \R^M, \aconst\in \R} \kua{\qs \sumi {\vi\over ||\vweight||}(2\wi - 1) - {|\aconst|\over ||\vweight||} }
\nonumber \\
&\Rightarrow& \hwem{Oracle bound-optimal rule:} \quad \vstar_i\propto 2\wi-1 \connect \astar= 0 .
\label{eqn:chooseVA}
\end{eqnarray}


The prediction function of  the oracle MAP rule is
$
\hyj= \sign\kua{\sumi \ln\kua{\upratio{\wi}} \cdot \zij + \ln\upratio{\pi}},
$
which is a hyperplane rule with weight $\vi^\ora= \ln\upratio{\wi}$ and $\aconst^\ora=\ln\upratio{\pi}$.

Since by Taylor expansion,
$
\ln{x\over 1-x}= (4x - 2) + O\kua{\kua{x-\inv{2}}^2},
$
we see that the weight of the oracle bound-optimal rule is the first order Taylor expansion of the oracle MAP rule.
Similar result and conclusion hold for the Dawid-Skene model as well,  but we omit them due to space limitations.

By observing that the oracle MAP rule is very close to
the oracle bound-optimal rule,
the oracle MAP rule approximately optimizes the upper bound of the mean error rate. This fact also indicates that our bound
is meaningful since the oracle MAP rule is the oracle Bayes classifier.

\subsection{Error rate bounds on one-step weighted majority voting}
\def \stepone{\hwem{ (Step 1) }}
\def \steptwo{\hwem{ (Step 2) }}
\def \stepthr{\hwem{ (Step 3) }}

Weighted majority voting (WMV) is a hyperplane rule with a shift constant $\aconst=0$, and if the weights of workers are the same, it will degenerate to majority voting.
From Section \ref{sec:bound_optimal}, we know that bound-optimal strategy of choosing weight is $\vi\propto 2(\wi -1)$. With this strategy, if we have an estimated $\wi$,  we can put more weights to the \lq\lq{}better\rq\rq{} workers and downplay the \lq\lq{}spammers\rq\rq{}( those workers with accuracy close to random guessing). This strategy can potentially improve the performance of majority vote and result in a better estimate for $\wi$.
This inspires us to design an iterative WMV method as follows:
\stepone Use majority voting to estimate labels, which are treated as ``golden standard".
\steptwo Use the current ``golden standard" to estimate the worker accuracy $\wi$ for all $i$ and set $\vi= 2\wi - 1$ for all $i$.
\stepthr Use the current weight $v$ in WMV to estimate updated ``golden standard", and then return to \steptwo until converge.

Empirically, this iterative WMV method converges fast. But it also suffers from the local optimal trap as EM does, and is generally  hard to analyze its error rate. However, we are able to obtain the error rate bound in the next theorem for a ``naive" version of it -- \emph{one-step WMV} (osWMV), which executes \stepone to \stepthr only once (i.e., without returning to \steptwo after \stepthr).


\begin{thm}\label{thm:mseWMVBound}

Under the \sds model, with label sampling probability $q=1$, let $\yjwmv$ be the label predicted by one-step WMV for the $j$th item,  if $\wbar \geq \inv{2}+\inv{M}+\sqrt{\frac{(M-1)\ln2}{2M^2}}$, the mean error rate of one-step weighted majority voting will be:
\vspace{-2mm}
\begin{eqnarray} \label{eqn:oswmv}
\MERwmv \leq \finalUpBound,
\end{eqnarray}
\vspace{-2mm}
where $\pdiv= \sqrt{\inv{M}\sumi (\wpi-\inv{2})^2}$ and  $\seta= 2\halfseta$
\end{thm}
The proof of this theorem is deferred to the supplementary. It is non-trivial to prove since the dependency among the weights and labels makes it hard to apply the concentration approach used in proving the previous results. Instead, a martingale-difference concentration bound has to be used.

\hwem{Remarks.}
(1) In the exponent of the bound, there are several important factors: $\pdiv$ represents how far the accuracies of workers are  away from random guessing, and it is a constant smaller than 1;  $\seta$ will be close to $0$ given a reasonable $M$.
(2) The condition on $\wbar$ requires that $\wbar - \inv{2}$ is $\Omega(M^{-0.5})$, which is easier to satisfy with $M$ large if the average accuracy in the crowd population is better than random guessing. This condition ensures majority voting  approximating  true labels, thus with  more items labeled, we can get a better estimate about the workers\rq{} accuracies, then the one-step WMV will improve the performance with better weights.
(3) We count now on how $M$ and $N$ affect the bound but defer the formal mathematical analysis to the supplementary after proving the theorem:
first, when both $M$ and $N$ increase but $\frac{M}{N}= r $ is a constant or decreases, the error rate bound  decreases. This makes sense because with the number of items labeled per worker increasing,   $\pih$ will be more  accurate, the weights will be closer to oracle bound optimal rule.
Second, when $M$ is fixed and $N$ increases, i.e., the number of items labeled increases, the upper bound on the error rate decreases.
Third, when $N$ is fixed and $M$ increases,the bound decreases when $M < \sqrt{N}$ and then increases when $M$ beyond $\sqrt{N}$.
Intuitively, when $M$  is larger than $N$ and $M$ increases,  the fluctuation of prediction score function $\sumi(2\pih -1)\zij$, where $\pih$ is the estimated accuracy of $i$th worker, will be large. This increases the chance to make more prediction errors. When $M$ is reasonably small (compared with $N$) but it increases, i.e., with more people labeling each item, the accuracy of majority vote will be improved according to Corollary \ref{res:oneCoin_MER}
, then the gain on the accuracy of estimate $\pih$ leads the weights in one-step WMV to be closer to oracle bound-optimal rule.

\section{Experiments}\label{sec:experiment}

In this section, we present numerical experiments on simulated data for comparing the EM-MAP rule with the oracle MAP rule. Meanwhile, we  plug the parameters estimated by EM into the bound of the oracle MAP rule in Theorem \ref{res:msr_oracleMAP}, which we call \emph{MAP plugin bound}. Note that the MAP plugin bound is not a true bound but an estimated one.
We also compare the EM-MAP with majority voting and MAP plugin bound on real-world data when oracle MAP is not available.  Furthermore, we simulate the one-step WMV and MV, then compare their bounds.

\begin{figure}[!bhtp]
\begin{center}
\begin{tabular}{ccc}
\includegraphics[width=0.33\columnwidth]{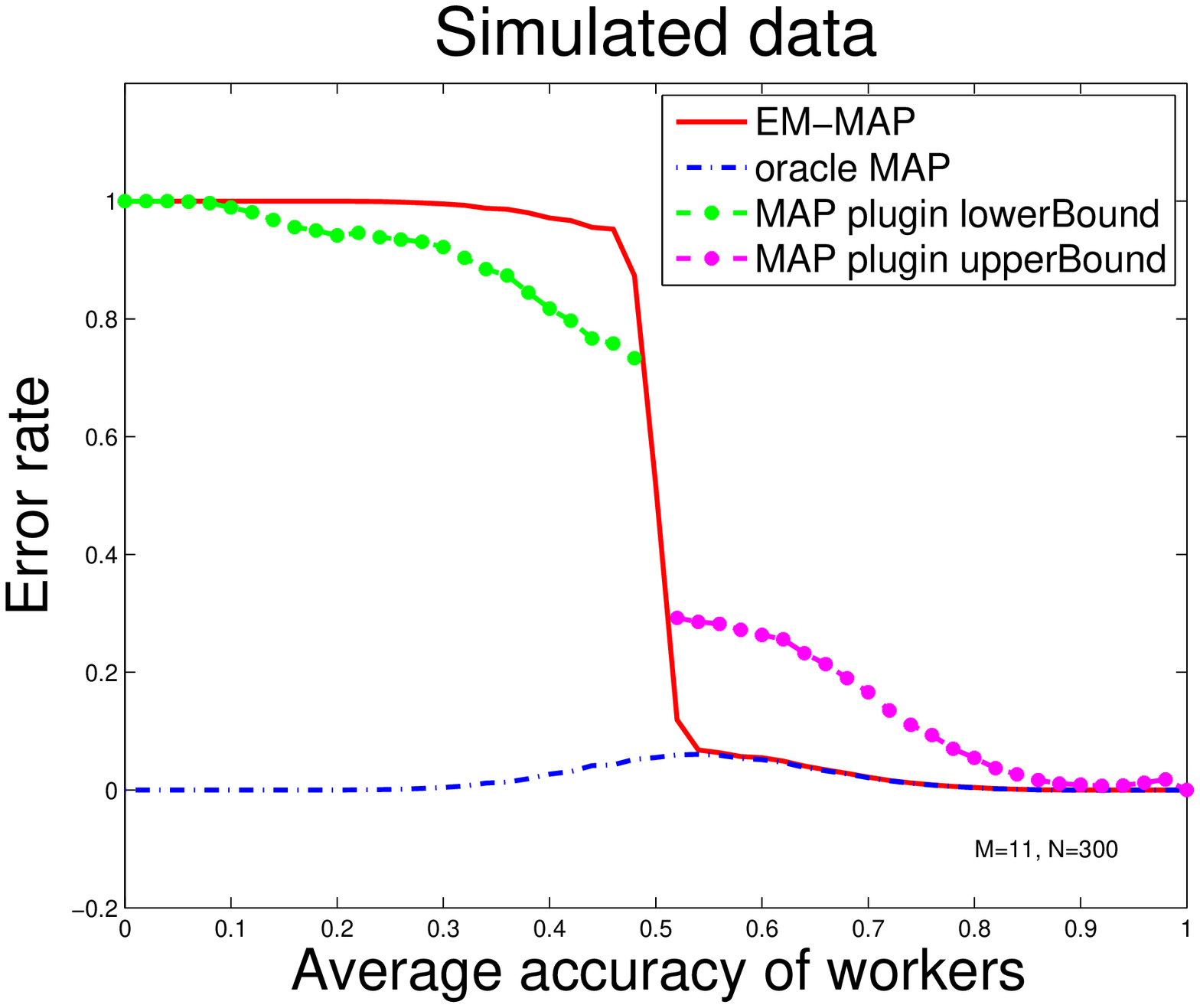} &
\includegraphics[width=0.33\columnwidth]{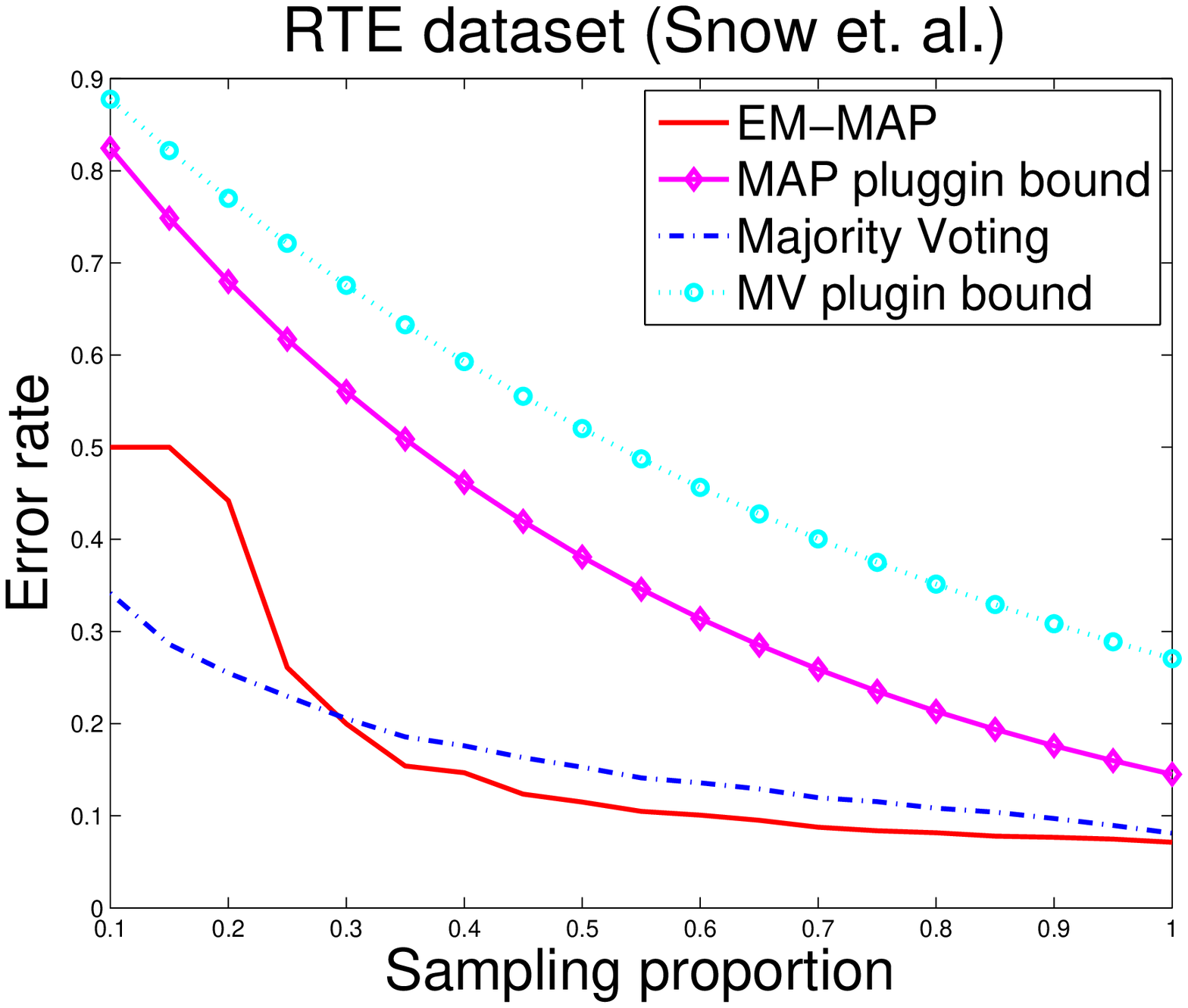}
&
\includegraphics[width=0.33\columnwidth]{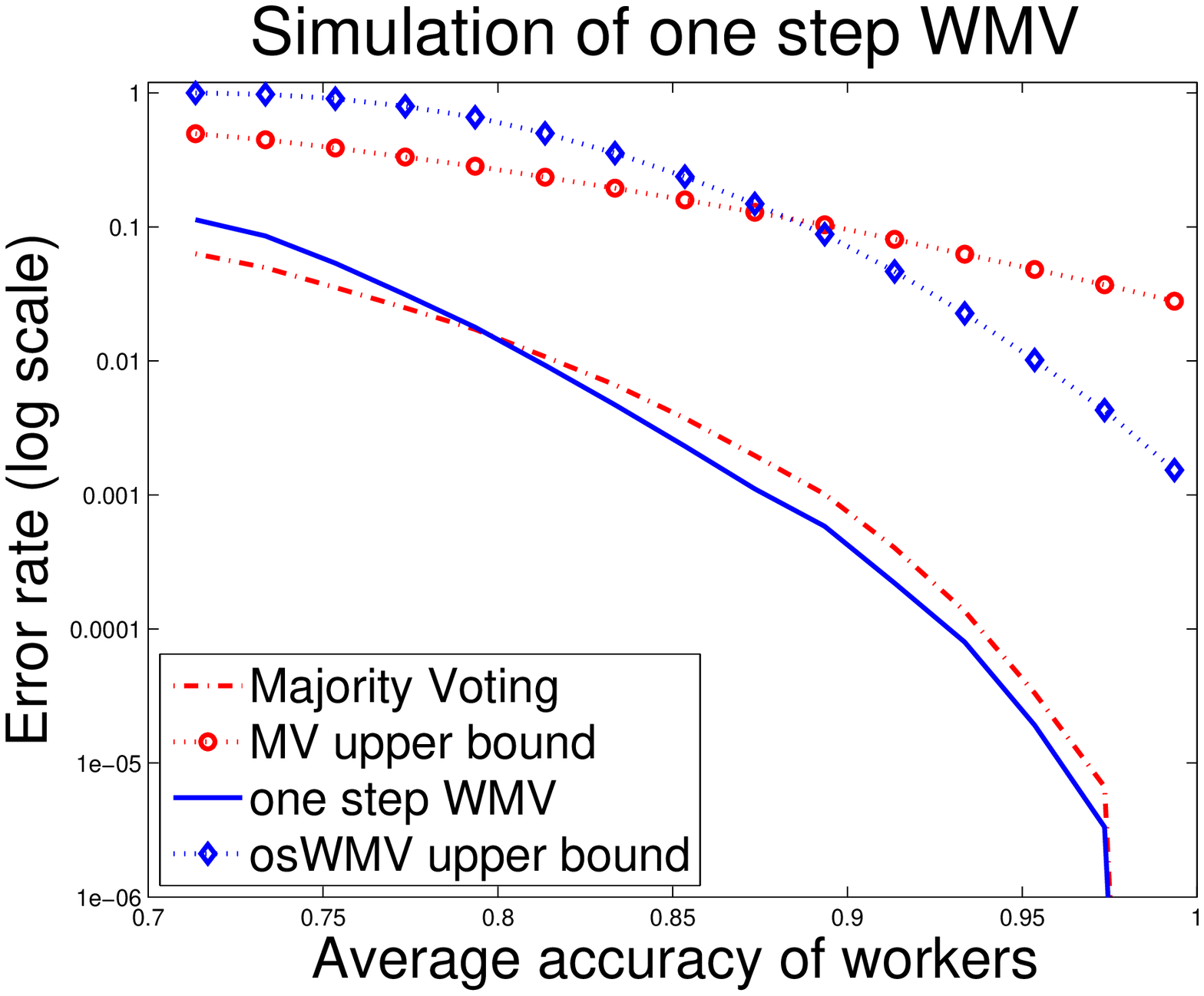}
\\
(a) & (b) & (c)
\end{tabular}
\caption{(a) Comparison of the EM-MAP rule and the oracle MAP rule by simulation. We plug the parameters estimated by EM into the oracle MAP bound and plot the plugin bound.
(b) Using the Snow et al. RTE dataset \cite{Snow_emnlp08} to compare the mean error rate of EM-MAP, majority voting and their bounds by plugging in the parameters estimated by comparing to the ground truth.
(c) Comparison of one-step weighted majority voting (osWMV), majority voting (MV) and their bounds by simulation.}
\label{fig:results}
\end{center}
\end{figure}

\textbf{Simulated data. }
The simulation is run under the \sds model with a constant sampling probability $q=0.8$ (each worker has 80\% chance to label any item). We simulate 11 workers who label 300 items with half of them being in the positive class. The workers\rq{} accuracies are sampled from beta distribution \emph{Beta}$(a, b)$ with $b=2. $ We control $a$  such that the expected accuracy of the workers varies from 0 to 1 with a step size 0.02. 
 The error rates are displayed in Fig. \ref{fig:results}(a).  Each error rate is averaged over 100 random data generations.

Fig. \ref{fig:results}(a) shows that when the average accuracy of workers is better than random guessing ($\wbar>0.5$), the EM-MAP rule almost has the same error rate as the oracle MAP rule (the red curve in Fig. \ref{fig:results}.(a) merges with the blue curve when $\wbar> 0.5$). It is interesting that when $\wbar < 0.5$, the  error rate of the oracle MAP rule is close to 0, while the error rate of the EM-MAP rule is close to 1. This is because if all the workers have low labeling accuracy, the EM algorithm cannot recognize this since there is no ground truth. While the oracle knows the true accuracies of workers, it can simply flip the labels from the workers. The plugin bounds are obtained by plugging the estimated worker accuracies by EM into the oracle MAP bound in Theorem \ref{res:msr_oracleMAP}.
When $\wbar> 0.5$, we compute the MAP plugin upper bound , and the lower bound when $\wbar < 0.5$. Note that whether  $\wbar > 0.5$ or not is the only information which we need but it depends on the true parameters. In real-world practice, if we have high confidence that the average worker accuracy is better than random guessing, such as in \cite{Ho2013}, Ho et al. using small amount of items with ground truth labels to test how good the workers are, then whether $\wbar > 0.5$ or not can be determined.


\textbf{Real data.} Fig.\ref{fig:results}(b) shows the result of the EM-MAP rule and majority voting on a language processing dataset from Snow et al. \cite{Snow_emnlp08}. The dataset is collected by asking workers to perform recognizing textual entailment (RTE) task, i.e.,  for each question the worker is presented with two sentences and given a binary choice of whether the second hypothesis sentence can be inferred from the first. There are 164 workers and 800 sentence pairs with ground truth collected from experts. Totally, 8000 labels are collected and on average each worker labeled 8000/(164$\times$800)= 6.1\% of items.

We estimate each worker\rq{}s labeling probability ($\qs_i$ for the $i$th worker) through dividing the number of items they labeled by 800.  Then a control variable --- the sampling proportion $x$, is a probability we further sample from the available labels. For example, if $x=0.6$ and the $i$th worker labeled 20 items, then we further sample these 20 labels with \emph{Bernoulli(0.6)} for each one. If an item labeled by $i$th worker has not been selected by further sampling, we treat it as missing label. In this way, we can control the sampling vector $\qvec$ (see section \ref{sec:problem_formulate}) with varying $x$ from 0 to 1 (with step size 0.05 in simulation).  Note that when $x=1$, we use all the 8000 labels. For each $x$, we repeat sampling label matrix with \emph{Bernoulli}($x$) on each available label, running the EM-MAP rule and MV, computing the plugin bounds by plugging the estimated workers\rq{} accuracies (compared with ground truth) into the upper bound of oracle MAP rule and the MV upper bound under the \sds model, for 40 times. In the end, we average the error rate of the EM-MAP rule and MV, the MAP plugin upper bound and MV bound respectively.  From Fig. \ref{fig:results}(b), we can see that the MAP plugin bound approaches the error rate of the EM-MAP rule with $x$ increasing.

Fig. \ref{fig:results}.(c) shows the simulation of one-step WMV and the comparison of its bounds with majority voting. We simulate 15 workers and 3000 items. The way we simulate is the same as what we did in Fig. \ref{fig:results}(a) described above. The differences are that we let the average accuracy of workers start from the minimum $\wbar$ required in Theorem \ref{thm:mseWMVBound} instead of 0 and we run one-step WMV, MV and compute their respective bounds like what we did for the EM-MAP rule.  We can see from Fig. \ref{fig:results} (c) that both the bound and the measured error rate exhibited ``cross" phenomenon --  majority voting is better than one-step WMV in the very beginning and then with average accuracy increasing, one-step WMV predominates MV because based on the ``well" estimated accuracy from MV, one-step WMV can weight each worker according to how good he/she is. 
Note that the error rate is in log-scale.
The reason that the tail of the error rate curves suddenly drop is because we only have finite $N$, thus error rate cannot be arbitrary close to 0.






\section{Conclusion}

In this paper, we have provided bounds on error rate
of general hyperplane labeling rules (in probability and in expectation) under the \ds crowdsourcing model that includes the \sds model as a special case.
Optimizing the mean error rate bound under the \ds model leads to a prediction rule that is a good approximation to the oracle MAP rule.  A data-driven WMV (one-step WMV) is proposed to approximate the oracle MAP with a theoretical guarantee on its error rate.
Through simulations under the \sds model (for simplicity) and simulations based on real data, we have three findings: (1) the EM-MAP rule is close to the oracle MAP rule with superior performance in terms of error rate, (2) the plugin bound for the oracle MAP rule is also applicable to the EM-MAP rule, and (3) the error rate of the one-step WMV is shown to  be bounded well by the theoretical bound.

To the best of our knowledge, this is the first extensive work on error rate bounds for general prediction rules under the practical \ds model for crowdsourcing. Our bounds are useful for explaining the effectiveness of different prediction rules/functions. In the future, we plan to extend our results to the multiple-labeling situation and explore other types of crowdsourcing applications.

%

%
\nocite{Sheng2008}
\nocite{Raykar_jmlr10}
\nocite{Jin_nips02}


\small{

\bibliography{arXiv_LYZ_crowdsourcing}

\begin{thebibliography}{10}

\bibitem{Dawid_JRSS79}
A.~P. Dawid and A.~M. Skene.
\newblock {Maximum Likelihood Estimation of Observer Error-Rates Using the EM
  Algorithm}.
\newblock {\em Journal of the Royal Statistical Society.}, 28(1):20--28, 1979.

\bibitem{Ho2013}
C.~Ho, S.~Jabbari, and J.~W. Vaughan.
\newblock {Adaptive Task Assignment for Crowdsourced Classification}.
\newblock In {\em ICML}, 2013.

\bibitem{Jin_nips02}
R.~Jin and Z.~Ghahramani.
\newblock {Learning with Multiple Labels}.
\newblock In {\em NIPS}, 2002.

\bibitem{Karger_NIPS2011}
D.~R. Karger, S.~Oh, and D.~Shah.
\newblock {Iterative learning for reliable crowdsourcing systems}.
\newblock In {\em NIPS}, 2011.

\bibitem{Liu2012}
Q.~Liu, J.~Peng, and A.~Ihler.
\newblock {Variational Inference for Crowdsourcing}.
\newblock In {\em NIPS}, 2012.

\bibitem{Raykar_jmlr10}
V.~C. Raykar, S.~Yu, L.~H. Zhao, C.~Florin, L.~Bogoni, and L.~Moy.
\newblock {Learning From Crowds}.
\newblock {\em Journal of Machine Learning Research}, 11:1297--1322, 2010.

\bibitem{Sheng2008}
C.~S. Sheng and F.~Provost.
\newblock {Get Another Label? Improving Data Quality and Data Mining Using
  Multiple, Noisy Labelers Categories and Subject Descriptors}.
\newblock {\em SIGKDD}, pages 614--622, 2008.

\bibitem{Smyth1995}
P.~Smyth, U.~Fayyad, M.~Burl, P.~Perona, and P.~Baldi.
\newblock {Inferring Ground Truth from Subjective Labelling of Venus Images}.
\newblock In {\em NIPS}, 1995.

\bibitem{Snow_emnlp08}
R.~Snow, B.~O. Connor, D.~Jurafsky, and A.~Y. Ng.
\newblock {Cheap and Fast - But is it Good ? Evaluating Non-Expert Annotations
  for Natural Language Tasks}.
\newblock {\em EMNLP}, 2008.

\bibitem{Welinder_nips10}
P.~Welinder, S.~Branson, S.~Belongie, and P.~Perona.
\newblock {The Multidimensional Wisdom of Crowds}.
\newblock In {\em NIPS}, 2010.

\bibitem{Whitehill_nips09}
J.~Whitehill, P.~Ruvolo, T.~Wu, J.~Bergsma, and J.~Movellan.
\newblock {Whose Vote Should Count More : Optimal Integration of Labels from
  Labelers of Unknown Expertise}.
\newblock In {\em NIPS}, 2009.

\bibitem{Yan_icml10}
Y.~Yan, R.~Rosales, G.~Fung, M.~Schmidt, G.~Hermosillo, L.~Bogoni, L.~Moy, and
  J.~G. Dy.
\newblock {Modeling annotator expertise : Learning when everybody knows a bit
  of something}.
\newblock In {\em ICML}, volume~9, pages 932--939, 2010.

\bibitem{Zhou2012}
D.~Zhou, J.~Platt, S.~Basu, and Y.~Mao.
\newblock {Learning from the Wisdom of Crowds by Minimax Entropy}.
\newblock In {\em NIPS}, 2012.

\end{thebibliography}
\bibliographystyle{plain}

}

\section*{Acknowledgements}
We thank Riddhipratim Basu and Qiang Liu for helpful discussions.


\begin{supplement}[id=suppA]
  \sname{Supplement A}
  \stitle{Missing proofs in this paper}
  \slink[url]{http://www.stat.berkeley.edu/$\sim$hwli/SupplementaryCrowdsourcing.pdf}
  \sdescription{ We have put all the missing proofs in the supplementary file, which can be downloaded from the link above.}
\end{supplement}

\end{document}